\def\year{2019}\relax
\documentclass[letterpaper]{article} 
\usepackage{aaai19}  
\usepackage{times}  
\usepackage{helvet}  
\usepackage{courier}  
\usepackage{url}  
\usepackage{graphicx}  
\usepackage{amsmath}
\usepackage{multirow}

\title{Learning Controllable Disentangled Representations with \\Decorrelation Regularization}
\author{Zengjie Song$^1$, Oluwasanmi Koyejo$^2$, Jiangshe Zhang$^1$\\
	$^1$School of Mathematics and Statistics, Xi'an Jiaotong University\thanks{This work was done while Zengjie Song was a visiting PhD student at University of Illinois at Urbana-Champaign.}, Xi'an, Shaanxi 710049, China\\
	$^2$Department of Computer Science, University of Illinois at Urbana-Champaign, Urbana, Illinois 61801, USA\\
    \texttt{zjsong@hotmail.com}, \texttt{sanmi@illinois.edu}, \texttt{jszhang@mail.xjtu.edu.cn}
}

\begin{document}
\maketitle

\begin{abstract}
A crucial problem in learning disentangled image representations is controlling the degree of disentanglement during image editing, while preserving the identity of objects. In this work, we propose a simple yet effective model with the encoder-decoder architecture to address this challenge. To encourage disentanglement, we devise a distance covariance based decorrelation regularization. Further, for the reconstruction step, our model leverages a soft target representation combined with the latent image code. By exploiting the real-valued space of the soft target representations, we are able to synthesize novel images with the designated properties. We also design a classification based protocol to quantitatively evaluate the disentanglement strength of our model. Experimental results show that the proposed model competently disentangles factors of variation, and is able to manipulate face images to synthesize the desired attributes.
\end{abstract}

\section{Introduction}
One of the long-standing challenges in machine learning community is to learn interpretable and robust representations of sensory data. Disentangling the hidden factors of variation provides a possible approach to overcome such a challenge \cite{Bengio13,Ridgeway16}. In a disentangled (or factorial) representation, the generative factors correspond to independent subsets of the latent dimensions, such that changing a single factor only causes a change in the single latent units.

Recently, many advances have been made in this direction. Those proposed models can be categorized into three prominent groups: 1) Generative Adversarial Network (GAN) \cite{Goodfellow14} based models \cite{Mirza14,Perarnau16,Donahue18}, 2) Autoencoder (AE) based models \cite{Kingma14a,Kingma14b,Bouchacourt18}, and 3) integrations of GAN's and AE's \cite{Larsen16,Makhzani16,Dumoulin17,Engel18}. It has been observed that GAN-based models can usually synthesize high-fidelity images such as faces and natural scenes, however, GAN's often suffer from unstable training and low sample diversity \cite{Salimans16}. On the other hand, AE-based models easily get stuck in generating blurry results \cite{Engel18}, yet fortunately do not seem to suffer from unstable training. Finally, hybrids of these two types of models try to implement a tradeoff between their strengths and weaknesses.

While the aforementioned models can produce compelling performance on manipulating images, they mainly focus on whether the model can generate images with or without attributes of interest, rather than the disentanglement degree, i.e., controlling the attribute intensity during image editing. However, a more subtle manipulation of images is often most useful in practice. For example, given a face image, one may desire not only to synthesize a new smiling face, but also to synthesize a sequence of faces with expressions varying from no smile to toothy smile. This property could allow several potential applications, such as automatic face image editing and image color rendering \cite{Lample17}.

Towards the goal of controlling disentanglement, we must still solve the problem of how to incorporate the designated attributes into the original image without changing other attribute information. In fact, this problem has also appeared in many previous works \cite{Cheung15,Kulkarni15,Chen16,Higgins17,Ma18,Kumar18}. For example, as observed in Figure 1 in \cite{Higgins17}, adding a fringe to a face concurrently leads to the visually perceptible change of skin color. Therefore, developing a flexible model that can modify designated attributes to various degrees, without destroying other characteristics is still an open but challenging task.

\begin{figure}[t]
	\centering
	\includegraphics[width=0.47\textwidth]{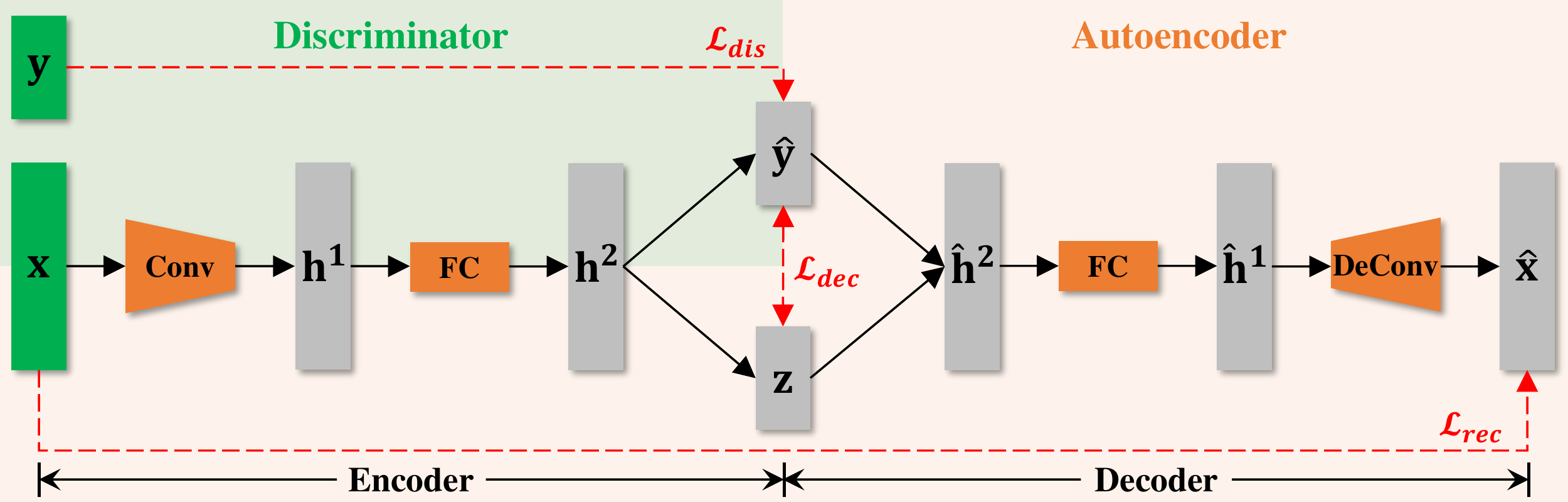}
	\caption{Network architecture of the proposed mddAE model. The symbols $\mathcal{L}_{rec}$, $\mathcal{L}_{dis}$, and $\mathcal{L}_{dec}$ indicate the reconstruction, discriminative, and decorrelation losses, respectively. See Section \ref{sec:Model} for a detailed definition.}\label{fig:model_structure}
\end{figure}

In this paper, we present mddAE (Autoencoder with manageable disentanglement degrees), a simple yet effective model that can be used to edit images while controlling the degree of disentanglement. As shown in Figure \ref{fig:model_structure}, by adding a discriminator to the basic AE, the mddAE can learn a soft target representation containing class or attribute-related knowledge. To facilitate disentanglement, we devise a novel decorrelation regularization based on the distance covariance (dCov) \cite{Szekely07}, which encourages the latent representation $\mathbf{z}$ to contain the information different from those in $\mathbf{\hat{y}}$. The ability of mddAE to control disentanglement degrees comes from the joint effect of the soft target representation and the decorrelation regularization. We assumption that, after performing decorrelation on these two representations, the value of the soft target representation should implicitly indicate how much attribute information is included in input image. To this end, we replace the discrete label, as used in many existing works \cite{Mirza14,Cheung15,Perarnau16,Makhzani16}, with the soft target representation, and then feed it along with the latent representation to the decoder for training. By doing so, the soft target representation learns to merge the reconstruction information with the decorrelation information in a simple and flexible manner, and thus improving the facticity of the synthesized images.

In summary, we highlight our contributions as follows.
\begin{itemize}
	\item We propose the mddAE model to learn disentangled representations, and further use it to edit images while controlling the degree of disentanglement.
	\item We devise a new decorrelation regularization based on the distance covariance, which can be viewed as an alternative to the cross covariance regularization. In particular, when evaluated on the representation disentangling task, our model equipped with the cross covariance regularization also outperforms an existing model that uses the same regularization \cite{Cheung15}.
	\item We design an evaluation protocol to quantitatively compare the disentanglement strength of our model. To our best knowledge, this is the first work that leverages classification to analyze how the effect of representation scales with disentanglement performance.
\end{itemize}

\section{Related Work}

\subsection{Learning Disentangled Representations}

\textbf{GAN-based models} \ The original GAN model \cite{Goodfellow14} does not show any apparent disentanglement properties, however, subsequent works have extended GANs. \citeauthor{Perarnau16} (2016) introduce an encoder in the conditional GAN model \cite{Mirza14}, and thus implement image editing by changing the conditional information inferred from the real image. \citeauthor{Donahue18} (2018) propose the semantically decomposed GANs, which learn to decompose the latent code into an identity-related portion and observation-related portion, thus modifying face images by varying the observation vector. By coupling two GANs together, the DiscoGAN model \cite{Kim17} leverages the cross-domain relations to perform the facial attribute conversion task. While GAN-based models are popular due to their superiority in generating high-fidelity images, they usually suffer from training instability and ``model collapse'' problems as demonstrated in \cite{Salimans16}.

\textbf{AE-based models} \ Alternatively, the AE-based models have empirically demonstrated the ability to synthesize diverse novel images. In fact, the vanilla Variational Autoencoder (VAE) \cite{Kingma14a,Rezende14} has been shown to learn disentangled representations, but with limited disentanglement ability on simple datasets such as FreyFaces or MNIST \cite{Kingma14a}. \citeauthor{Kingma14b} (2014) and \citeauthor{Siddharth17} (2017) formulate semi-supervised learning in the context of VAEs, and using label information to achieve conditional generation. Another VAE-based model is the multi-level VAE \cite{Bouchacourt18}, designed for learning disentangled factors of grouped data. However, we note that AEs and VAEs are prone to produce blurry reconstructions \cite{Engel18}.

\textbf{Combined GAN and AE models} \ A natural way to alleviate above mentioned problems is to integrate GAN and AE, and thus leveraging both models' strengths in a complementary manner. \citeauthor{Larsen16} (2016) propose the VAE/GAN model, which views the VAE decoder and the GAN generator as the same mapping and uses the GAN discriminator to measure sample similarity. In addition to apply the GAN in image space, it's also possible to explore the adversarial training in latent space of AEs. The main point of such models is making the discriminator indistinguishable 1) between the aggregated posterior of the latent variable and an arbitrary prior \cite{Makhzani16}; 2) or between samples in latent space and encoded data (rather than prior samples) \cite{Engel18}; 3) or between joint samples of the data and the corresponding latent variable from the encoder and joint samples from the decoder \cite{Dumoulin17}.

\subsection{Controlling Disentanglement}
The InfoGAN \cite{Chen16} can change a subset of facial attributes by manipulating the learned categorical codes, but with no conspicuous visual difference among generated images, such as the ``Hair style'' variation shown in Figure 6 in \cite{Chen16}. Focusing on person image generation, \citeauthor{Ma18} (2018) use an adversarial network to learn mappings from Gaussian noise to the embedding feature space, and thus providing more control over the foreground, background, and pose information of the input image. \citeauthor{Denton17} (2017) and \citeauthor{Tulyakov18} (2018) employ adversarial training to learn disentangled representations of video, and generate a sequence of video frames with the same content but different motions and vice versa.

\citeauthor{Higgins17} (2017) and \citeauthor{Kumar18} (2018) extend VAE to learn controllable disentangled factors, implemented by putting implicit independence constraints on the approximate posterior over latent variables. However, in these models, the change of one attribute (e.g., smiling) may simultaneously result in changes of other attributes (e.g., hairstyle or azimuth) as observed in \cite{Kumar18}. \citeauthor{Kulkarni15} (2015) achieve disentanglement based on a special training scheme, where pairs of rendered images that differ only in one factor of variation are provided. Another VAE-like model \cite{Hou17} utilizes the vector arithmetic technique \cite{Mikolov13} to control attribute intensities. Both of these models yield impressive results, but they rely on either structured training data \cite{Kulkarni15} or precomputed attribute-specific representations \cite{Hou17}, which is generally not flexible to address more complicated image generation tasks (e.g., manipulating a number of face images with multiple facial attributes).

By constraining the latent variables to be invariant to attributes, the basic AE model can also be extended to address disentanglement. The FadNet \cite{Lample17} applies an adversarial-like process to approach such a goal, while \citeauthor{Cheung15} (2015) use the cross covariance (XCov) regularization instead. To control the attribute intensity, these models consider each attribute as a continuous variable during image editing, which is similar to managing the soft target representation in our model. However, training the FadNet relies on several delicate-designed tricks (e.g., weighting effects of discriminator cost with a hyperparameter scheduling process), which may need to carefully reset as applied to different datasets. By contrast, our model is simpler\textemdash{}thus easier to use for real world applications (e.g., utilizing constant hyperparameters), and meanwhile requires no adversarial training. On the other hand, compared with the XCov regularization, the distance covariance (dCov) we use for disentanglement encourages statistical independence rather than non-correlation between variables, leading to a stronger disentanglement ability. More importantly, instead of feeding decoder with the discrete label during training in \cite{Cheung15}, our model leverages a continuous target representation to facilitate reconstruction, which results in simpler fusion of the original identity and the modified attribute with different variation degrees.

\section{Model}\label{sec:Model}
In this section, we first give a brief description of the network architecture, then discuss the model loss and various disentanglement-inducing regularizers. Finally, the method to manipulate images with controllable disentanglement is illustrated in two application cases.

\subsection{Network Architecture and Total Loss}
As shown in Figure \ref{fig:model_structure}, the network architecture of mddAE is established based on the encoder-decoder framework, with an embedded discriminator featuring in the representation learning. Specifically, the middle-layer representation is still inferred from the input image $\mathbf{x}$ by encoder, but it is divided into two parts: the \emph{soft target representation} $\mathbf{\hat{y}}$ and the \emph{latent representation} $\mathbf{z}$. Through the discriminator, we inject the class or attribute knowledge into $\mathbf{\hat{y}}$, which is designed to take the form of probabilistic representation \emph{during training}. The decoder is then fed with these two representations to reconstruct input image.

In addition to the aforementioned symbols\footnote{Without loss of clarity, we may omit the subscript ``$n$'' used to index samples henceforth.}, let $N$ be the mini-batch size, $\mathbf{\hat{x}}$ be the reconstruction of input $\mathbf{x}$, and $\mathbf{\hat{Y}}$ and $\mathbf{Z}$ correspond to the mini-batch soft target representation $\mathbf{\hat{y}}$ and the latent representation $\mathbf{z}$, respectively. The total loss function of the mddAE model is:
\begin{align}\label{eqn:loss_total}
  \mathcal{L} = \ &\frac{1}{N}\sum_{n}^{N}\mathcal{L}_{rec}(\mathbf{x}_{n}, \mathbf{\hat{x}}_{n}) + \beta\frac{1}{N}\sum_{n}^{N}\mathcal{L}_{dis}(\mathbf{y}_{n}, \mathbf{\hat{y}}_{n}) \nonumber \\
                  &+ \gamma\mathcal{L}_{dec}(\mathbf{\hat{Y}}, \mathbf{Z})
\end{align}
where $\mathcal{L}_{rec}$, $\mathcal{L}_{dis}$, and $\mathcal{L}_{dec}$ denote the \emph{reconstruction loss}, \emph{discriminative loss}, and \emph{decorrelation loss}, respectively; $\beta$ and $\gamma$ are two constant weights. Briefly, minimizing the reconstruction loss enables the whole model to reconstruct images as accurately as possible. And minimizing the discriminative loss helps the encoder to extract class or attribute-related information. The decorrelation loss (or regularization) facilitates the latent representation $\mathbf{z}$ to be different from the soft target representation $\mathbf{\hat{y}}$, and thus forcing the model to retain all information independent of the class or attribute in $\mathbf{z}$. Next, we formulate these three losses and clarify their roles in learning disentangled representations.

\subsection{Reconstruction Loss $\mathcal{L}_{rec}$}
The reconstruction loss $\mathcal{L}_{rec}(\mathbf{x}, \mathbf{\hat{x}})$ measures the difference between original image $\mathbf{x}$ and its reconstruction $\mathbf{\hat{x}}$. A common choice is the Mean Squared Error (MSE), defined as
\begin{equation}\label{eqn:loss_mse}
\mathcal{L}_{mse}(\mathbf{x}, \mathbf{\hat{x}}) = \|\mathbf{x} - \mathbf{\hat{x}}\|_{2}^{2}
\end{equation}
where $\|\cdot\|_{2}^{2}$ represents the squared $l^{2}$-norm. The fundamental computation principle used in our model is fairly straightforward like the basic AE, that is, the soft target representation $\mathbf{\hat{y}}$ and the latent representation $\mathbf{z}$ are first computed by the encoder, and then utilized by the decoder to generate the reconstruction:
\begin{equation}\label{eqn:encoder_decoder}
\{\mathbf{\hat{y}}, \mathbf{z}\} = \text{Encoder}(\mathbf{x}; \phi), \quad \mathbf{\hat{x}} = \text{Decoder}(\mathbf{\hat{y}}, \mathbf{z}; \theta)
\end{equation}
where $\phi$ and $\theta$ denote the network parameters of encoder and decoder, respectively. Here we formulate the soft target representation $\mathbf{\hat{y}}$ in two application cases. 1) For the case of multiple classes with competition between them (e.g., a handwritten digit belongs to only one of the 10 classes), we use the softmax nonlinearity to compute each element of $\mathbf{\hat{y}}$:
\begin{equation}\label{eqn:y_hat_softmax}
\hat{y}_{i} = \frac{e^{-a_{i}}}{\sum_{j=1}^{C}e^{-a_{j}}}, i = 1, 2, \dots, C
\end{equation}
where $a_{i}$ indicates the input of the $i$-th representation unit, and $C$ is the number of all such units (i.e., the dimension of $\mathbf{\hat{y}}$). 2) For another case of multiple attributes where each attribute has binary classes (e.g., face images with or without smiling, eyeglasses, blond hair, etc.), we use the sigmoid nonlinearity to compute each element of $\mathbf{\hat{y}}$:
\begin{equation}\label{eqn:y_hat_sigmoid}
\hat{y}_{i} = \frac{1}{1 + e^{-a_{i}}}, i = 1, 2, \dots, C.
\end{equation}

To alleviate the problem of blurry reconstruction, we incorporate the well-studied multi-scale Structural Similarity (SSIM) index \cite{Wang03} to improve the perceptual quality of reconstructions. Based on this index, we obtain the following structural dissimilarity as an auxiliary reconstruction loss\footnote{For a detailed description on the multi-scale SSIM index, refer to \cite{Wang03}.}:
\begin{equation}\label{eqn:loss_dssim}
\mathcal{L}_{dssim}(\mathbf{x}, \mathbf{\hat{x}}) = \frac{1}{2}(1 - \text{SSIM}(\mathbf{x}, \mathbf{\hat{x}})).
\end{equation}
Therefore, the final reconstruction loss is:
\begin{equation}\label{eqn:loss_rec}
\mathcal{L}_{rec}(\mathbf{x}, \mathbf{\hat{x}}) = \mathcal{L}_{mse}(\mathbf{x}, \mathbf{\hat{x}}) + \alpha\mathcal{L}_{dssim}(\mathbf{x}, \mathbf{\hat{x}})
\end{equation}
where $\alpha$ is a constant.


\subsection{Discriminative Loss $\mathcal{L}_{dis}$}
We also extend the discriminative loss $\mathcal{L}_{dis}$ to the aforementioned two application cases. 1) For the first case which corresponds to a classification problem with $C$ classes, we employ the cross entropy between the discrete label $\mathbf{y}$ and the soft target representation $\mathbf{\hat{y}}$ as the discriminative loss:
\begin{equation}\label{eqn:loss_class_ce}
\mathcal{L}_{dis}(\mathbf{y}, \mathbf{\hat{y}}) = -\sum_{i=1}^{C}y_{i}\cdot\log\hat{y}_{i}
\end{equation}
where $\hat{y}_{i}$ is computed by the softmax nonlinearity in \eqref{eqn:y_hat_softmax}. 2) For the second case which actually corresponds to $C$ binary classification problems, we use the binary cross entropy as the discriminative loss:
\begin{equation}\label{eqn:loss_class_bce}
\mathcal{L}_{dis}(\mathbf{y}, \mathbf{\hat{y}}) = -\frac{1}{C}\sum_{i=1}^{C}[y_{i}\cdot\log\hat{y}_{i} + (1-y_{i})\cdot\log(1-\hat{y}_{i})]
\end{equation}
where $\hat{y}_{i}$ is computed by the sigmoid nonlinearity in \eqref{eqn:y_hat_sigmoid}.

\subsection{Decorrelation Loss $\mathcal{L}_{dec}$}
To encourage disentanglement, we propose to leverage the distance covariance (dCov) \cite{Szekely07} based regularization to learn the latent representation $\mathbf{z}$, which is expected to contain information different from class or attribute-related information in $\mathbf{\hat{y}}$.

Let $(\mathbf{\hat{y}}_{n}, \mathbf{z}_{n}), n = 1, 2, \dots, N$ be a statistical sample from a pair of soft target and latent random variables $(\mathbf{\hat{Y}}, \mathbf{Z})$. To obtain the decorrelation loss (or regularization), we first compute the $N$ by $N$ distance matrices $(a_{n,m})$ and $(b_{n,m})$ containing all pairwise distances:
\begin{align}
a_{n,m} &= \parallel \mathbf{\hat{y}}_{n} - \mathbf{\hat{y}}_{m}\parallel, \quad n,m = 1, 2, \dots, N,\\
b_{n,m} &= \parallel \mathbf{z}_{n} - \mathbf{z}_{m}\parallel, \quad n,m = 1, 2, \dots, N
\end{align}
where $\|\cdot\|$ denotes the Euclidean norm. Then take all doubly centered distances
\begin{align}
A_{n,m} &:= a_{n,m} - \bar{a}_{n\cdot} - \bar{a}_{\cdot m} + \bar{a}_{\cdot\cdot},\\
B_{n,m} &:= b_{n,m} - \bar{b}_{n\cdot} - \bar{b}_{\cdot m} + \bar{b}_{\cdot\cdot}
\end{align}
where $\bar{a}_{n\cdot}$ is the $n$-th row mean, $\bar{a}_{\cdot m}$ is the $m$-th column mean, and $\bar{a}_{\cdot\cdot}$ is the grand mean of the distance matrix of $\mathbf{\hat{Y}}$. The notation is similar for the $b$ values. Finally, the squared sample distance covariance, treated as our decorrelation loss, is simply the arithmetic average of the products $A_{n,m}B_{n,m}$:
\begin{equation}\label{eqn:loss_dec_dCov}
\mathcal{L}_{dec}(\mathbf{\hat{Y}}, \mathbf{Z}) = \text{dCov}^{2}(\mathbf{\hat{Y}}, \mathbf{Z}) = \frac{1}{N^{2}}\sum_{n=1}^{N}\sum_{m=1}^{N}A_{n,m}B_{n,m}.
\end{equation}

By comparison, \citeauthor{Cheung15} (2015) use the cross covariance (XCov) to facilitate disentanglement, which is given as
\begin{equation}\label{eqn:loss_dec_XCov}
\text{XCov}(\mathbf{\hat{Y}}, \mathbf{Z}) = \frac{1}{2}\sum_{i,j}[\frac{1}{N}\sum_{n=1}^{N}(\hat{y}_{n,i}-\bar{\hat{y}}_{i})(z_{n,j}-\bar{z}_{j})]^{2}
\end{equation}
where $\hat{y}_{n,i}$ indicates the $i$-th element of $\mathbf{\hat{y}}_{n}$, and $\bar{\hat{y}}_{i}$ is the mean of the $i$-th element across mini-batch samples. The notation is similar for $z_{n,j}$ and $\bar{z}_{j}$. Note that one of the most important difference between dCov and XCov is that, minimizing the dCov encourages the independence between two random variables \cite{Szekely07}, while minimizing the XCov encourages the non-correlation. To this end, the dCov should induce stronger disentanglement than XCov. Additionally, our model is also compatible with the XCov regularization, and replacing the $\text{dCov}^{2}$ with XCov in mddAE achieves improved disentanglement performance over the model in \cite{Cheung15}.

\subsection{Methods to Manipulate Images}\label{sec:method_manipulate}
At image editing time, the key operation is to modify the value of soft target representation $\mathbf{\hat{y}}$ accordingly. Due to the two potential application cases mentioned above, we give two related methods to perform image manipulation.

\begin{figure}[t]
	\centering
	\includegraphics[width=0.47\textwidth]{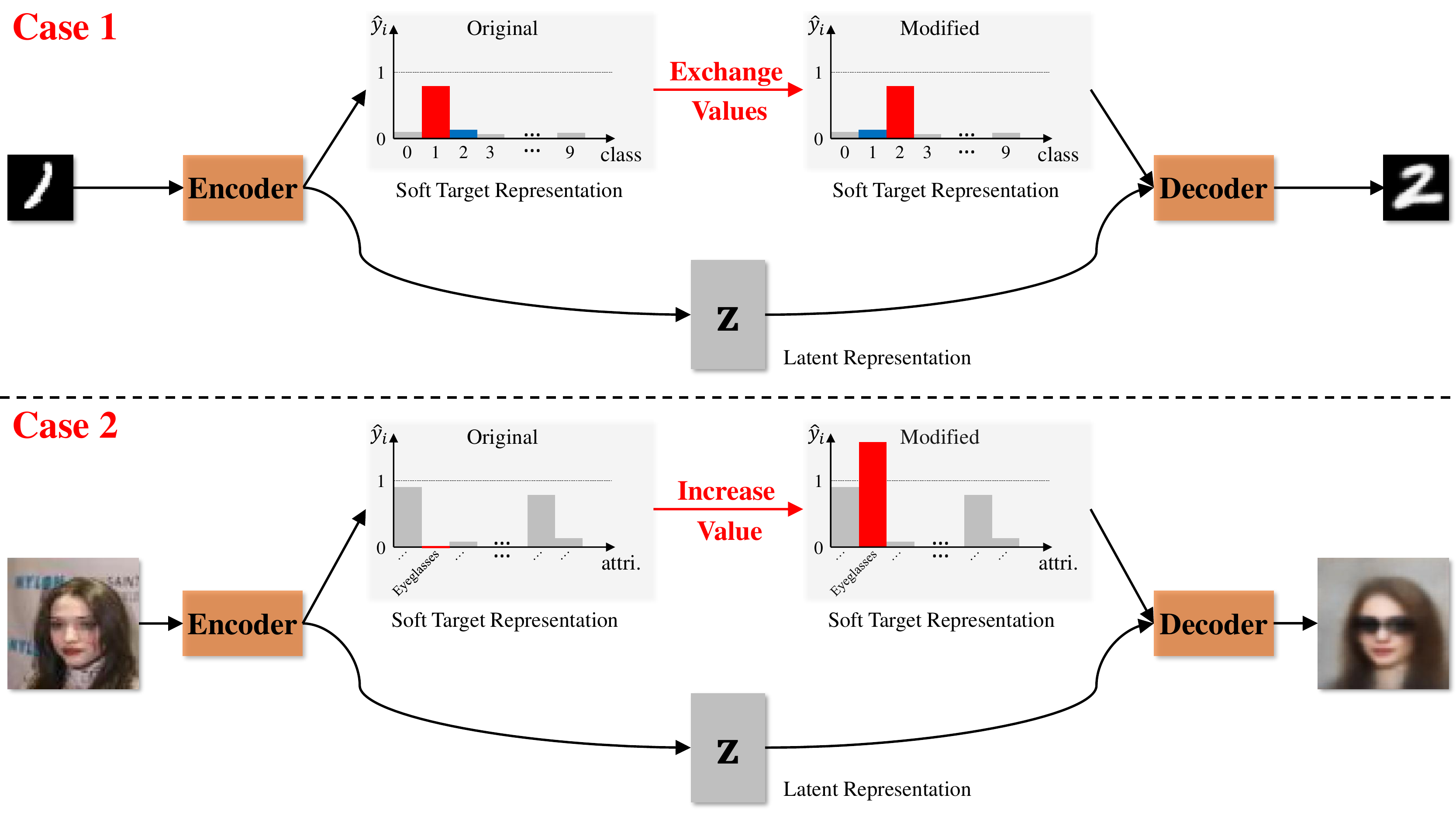}
	\caption{Manipulating images. The case 1 addresses the disentanglement task where the mutual exclusion exists among multiple classes. The case 2 is for the scenario where multiple attributes are independent of each other. The bigger version of this figure is provided in the supplementary material.}\label{fig:method_manipulate_images}
\end{figure}

In the first case, taking the handwritten digit as an example, we want to generate a new digit with the handwriting style designated by a given digit. To do this, as shown in Figure \ref{fig:method_manipulate_images}, we first employ the encoder to infer the soft target representation $\mathbf{\hat{y}}$ and the latent representation $\mathbf{z}$ of the digit ``1'' in boldface. Then we modify $\mathbf{\hat{y}}$ by exchanging the third element (corresponding to digit 2 class) and the maximum element (ideally corresponding to digit 1 class), while keeping remaining elements fixed. In this way, only two elements of $\mathbf{\hat{y}}$ at most are exchanged, and thus the representation structure with component summation of 1 is preserved completely. Finally, we feed the modified $\mathbf{\hat{y}}$ and the unchanged $\mathbf{z}$ to the decoder to generate the new digit ``2'' which is also in boldface.

In the second case, consider the face image as an example, and the goal is to synthesize a new face with the desired attribute and intensity while preserving the core identity. As we can see from Figure \ref{fig:method_manipulate_images}, the overall procedure is similar to the first case, but with a different modification of $\mathbf{\hat{y}}$. Specifically, in order to generate a new face with eyeglasses, we just replace the original (near) zero value corresponding to ``Eyeglasses'' attribute with the new value (e.g., 1.7) in $\mathbf{\hat{y}}$. Here we emphasize that \emph{during image editing}, the modified attribute value is not necessarily restricted in $[0, 1]$, and it can also take other real values greater than 1. By doing so, the soft target representation is able to cover a wide range that the network was never trained on and we will get meaningful generalization (see Section \ref{sec:disentangle_degree} for experimental evidences).

It's worth noting that we need no class or attribute labels during image editing, since we perform modification on the soft target representation inferred from input image, rather than on the original discrete label vector.

\section{Experiments}
In this section, we conduct three groups of experiments to evaluate the disentanglement performance of our mddAE model. First, we verify that the proposed decorrelation regularization and the image manipulation methods can disentangle factors of variation. Second, we explore the disentanglement strength of our model under various attribute intensities, so as to illustrate the ability of the mddAE to control the degree of disentanglement. Both of these two groups of experiments obtain qualitative results. Third, we leverage a classification based protocol to quantitatively compare the disentanglement strength of the mddAE. Additional results are provided in the supplementary material.

\subsection{Experimental Setup}
\textbf{Dataset} The evaluations are performed on two representative datasets. The first one is \textbf{MNIST} \cite{LeCun98}, which contains 70,000 grayscale handwritten digit images with $28\times28$ pixels for each and scaled to $[0, 1]$. We randomly split the dataset into 50,000 training, 10,000 validation, and 10,000 test samples, respectively. The discrete label has the one-hot vector form. The second dataset is \textbf{CelebA} \cite{Liu15}, which consists of 202,599 RGB face images of celebrities. For pre-processing, we resized all face images to $64\times64\times3$, and then the image values were normalized to $[-1, 1]$. We use $80\%$ images for training, $10\%$ for validation, and $10\%$ for test as used in several earlier works. Additionally, the discrete label is represented by the binary vector with dimension 40, where each dimension corresponds to one attribute with value 1 indicating containing this attribute and 0 not.

\textbf{Compared Models} The plain version of the mddAE, which includes no regularization and uses discrete label to help reconstruct images, is named disAE and treated as one of our baselines. Our model is inspired by the model of \citeauthor{Cheung15} (2015), which amounts to adding the cross covariance (XCov) regularization to disAE. Hence we symbolize this model with disAE-XCov and also use it as another baseline. While we focus on demonstrating the disentanglement ability of the proposed regularization and the image manipulation method based on the basic AE, it should be straightforward to extend these ideas to GAN and VAE-like models.

\textbf{Training Details} For all compared models, the encoder consists of convolution (Conv) layers followed by fully-connected (FC) layers, and the decoder is symmetric to encoder, but using deconvolution (DeConv, or the transposed convolution) \cite{Radford16} for the up-sampling. The network architecture details can be found in the supplementary material. Besides, we fix $\alpha=1$ and $\beta=1$ in all experiments. Based on the validation-set performance, the value of $\gamma$ is set to 5 across both datasets and experiments in the following Section \ref{sec:disentangle_ability} and \ref{sec:disentangle_degree}. All models are trained with the Adam optimizer \cite{Kingma15}, where we set learning rate $=1\text{e}-4$, $\beta_{1}=0.5$, $\beta_{2}=0.999$, and a batch size of 100 for MNIST, 128 for CelebA.

\begin{figure}[ht]
	\centering
	\includegraphics[width=0.23\textwidth]{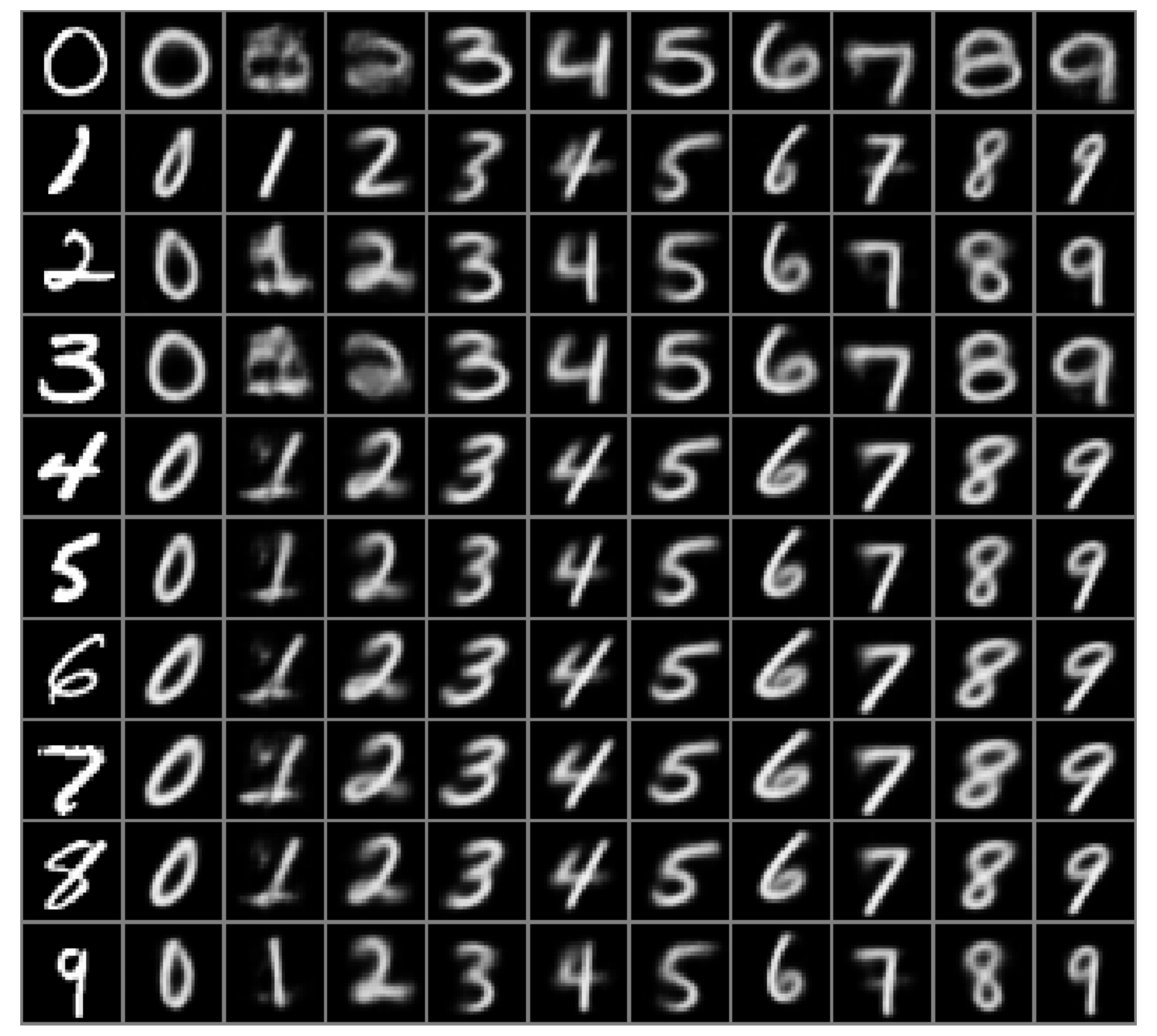}\hspace{0.15mm}
	\includegraphics[width=0.23\textwidth]{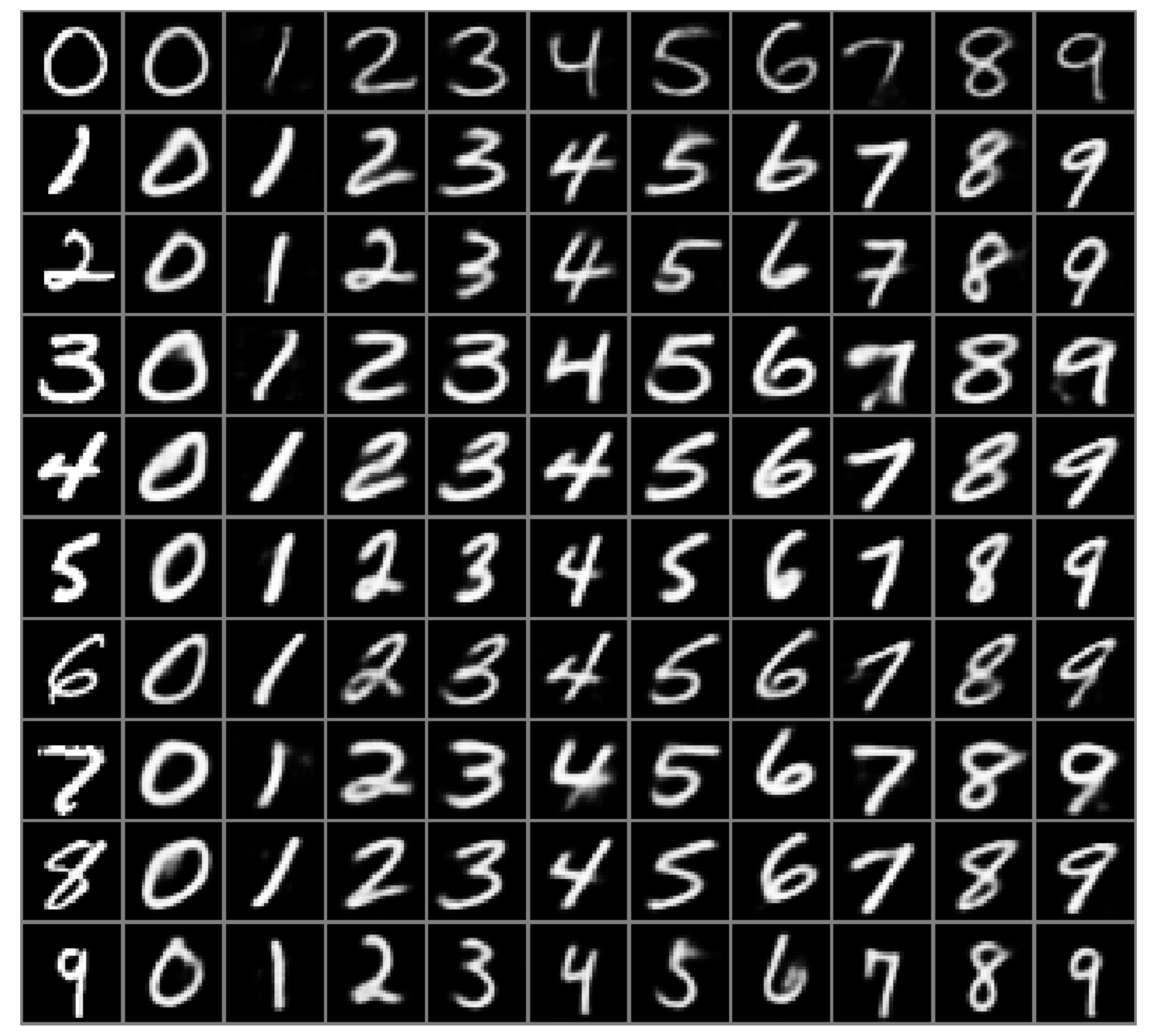}\\ \vspace{1.2mm}
	\includegraphics[width=0.23\textwidth]{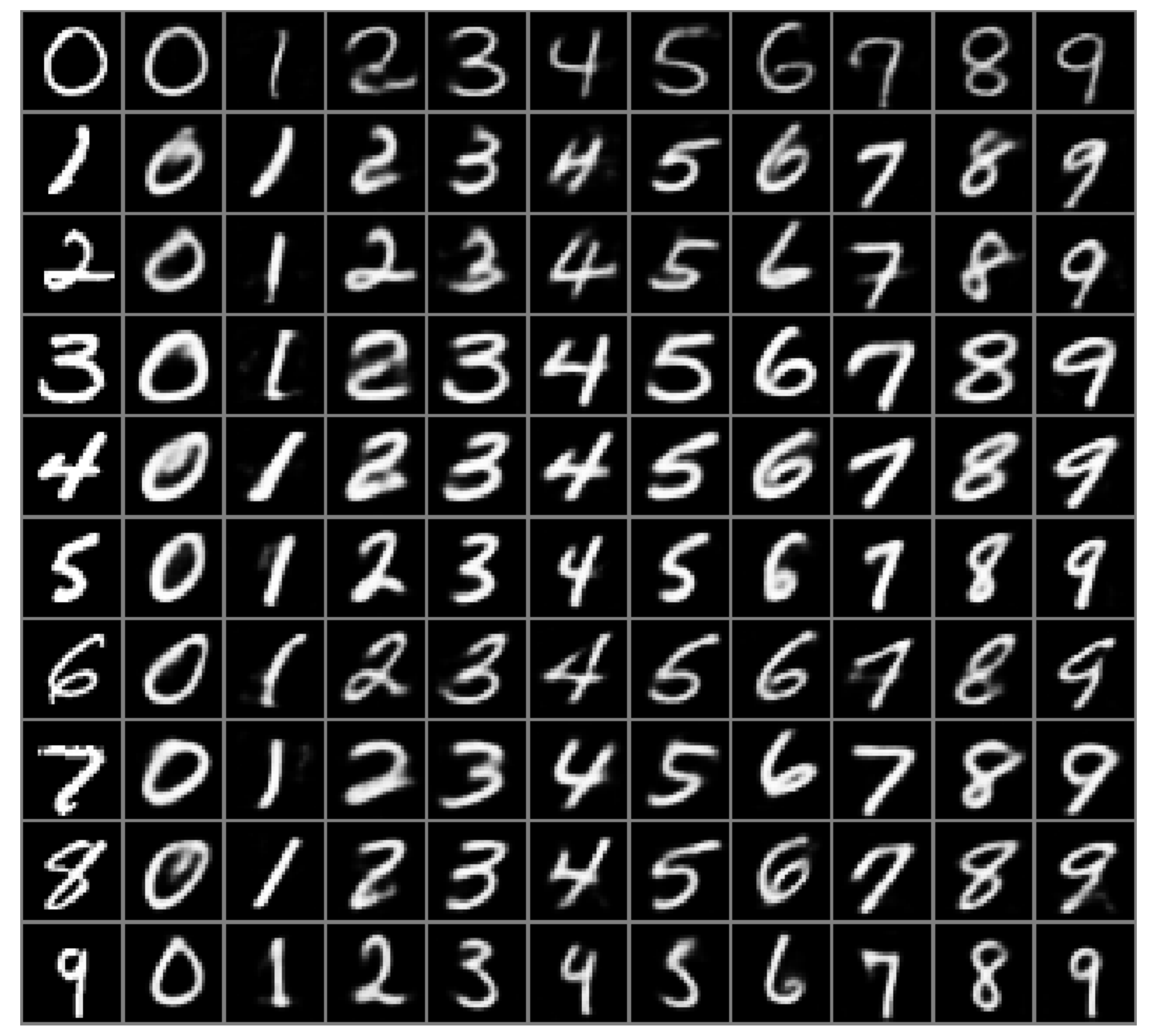}\hspace{0.15mm}
	\includegraphics[width=0.23\textwidth]{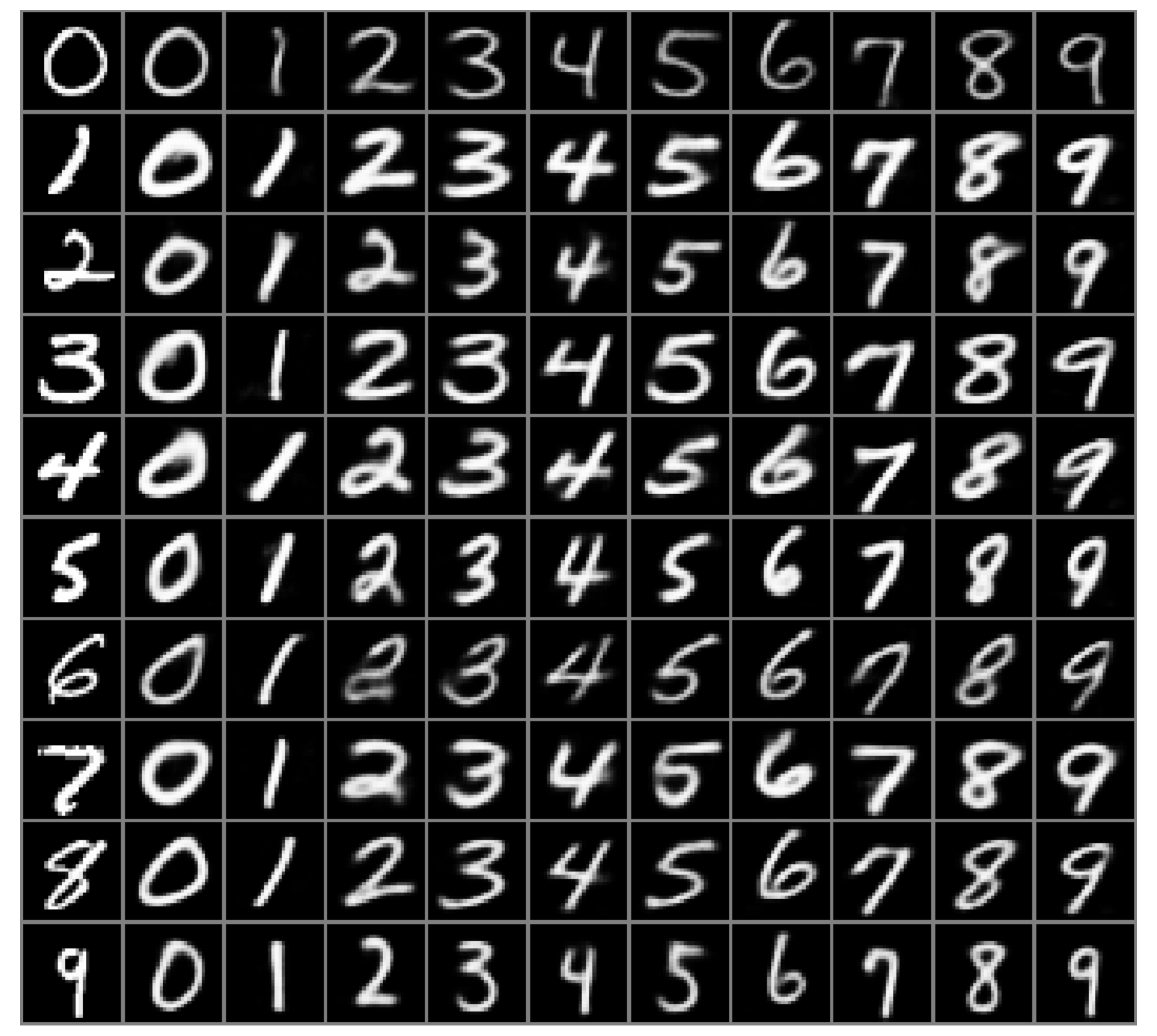}%
	\caption{Generated digits with different handwriting styles. \textbf{Baselines}: \textbf{disAE} (top-left) and \textbf{disAE-XCov} (top-right). \textbf{Ours}: \textbf{mddAE-XCov} (bottom-left) and \textbf{mddAE-dCov} (bottom-right). In each panel, the first column displays the MNIST test images, and the other columns show analogical fantasies of test images.}\label{fig:generate_digits}
\end{figure}
\begin{figure*}
	\centering
	\includegraphics[width=0.49\textwidth]{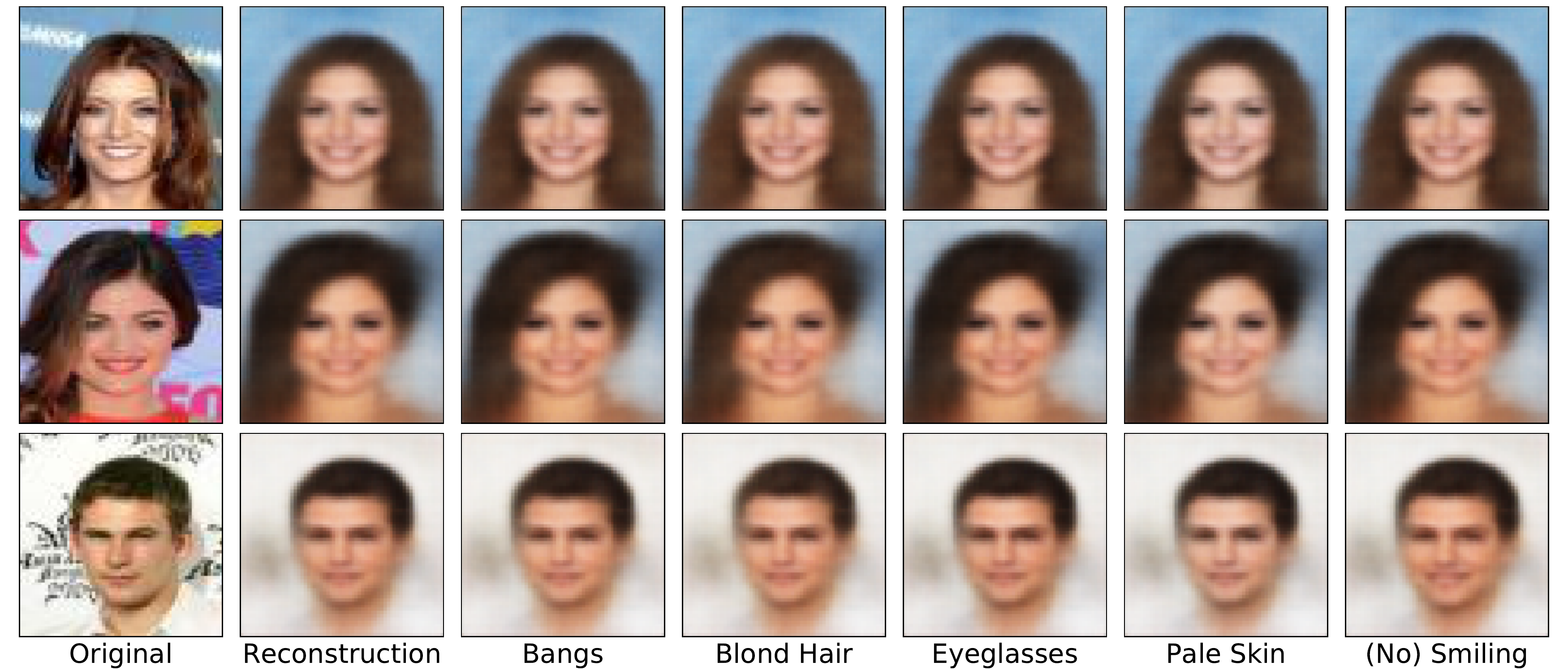}\hspace{0.2mm}
	\includegraphics[width=0.49\textwidth]{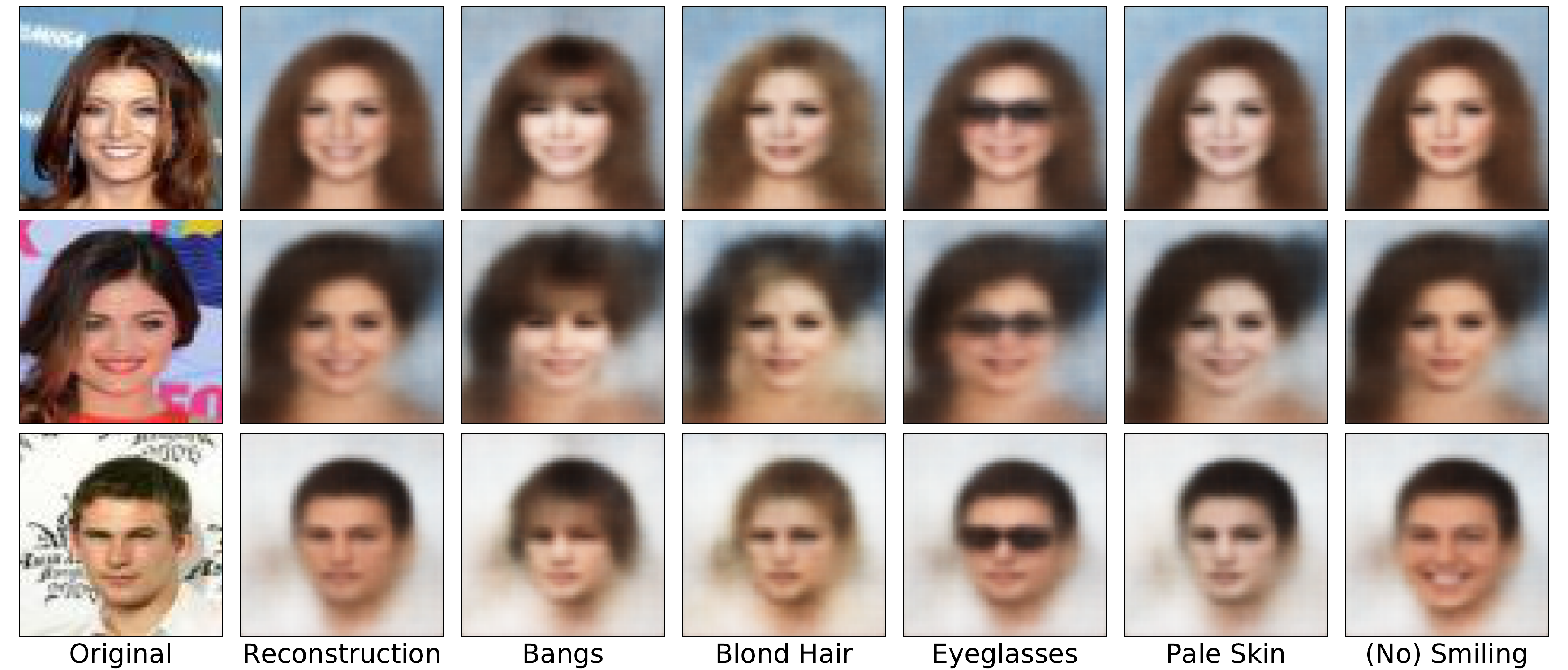}\\ \vspace{1.2mm}
	\includegraphics[width=0.49\textwidth]{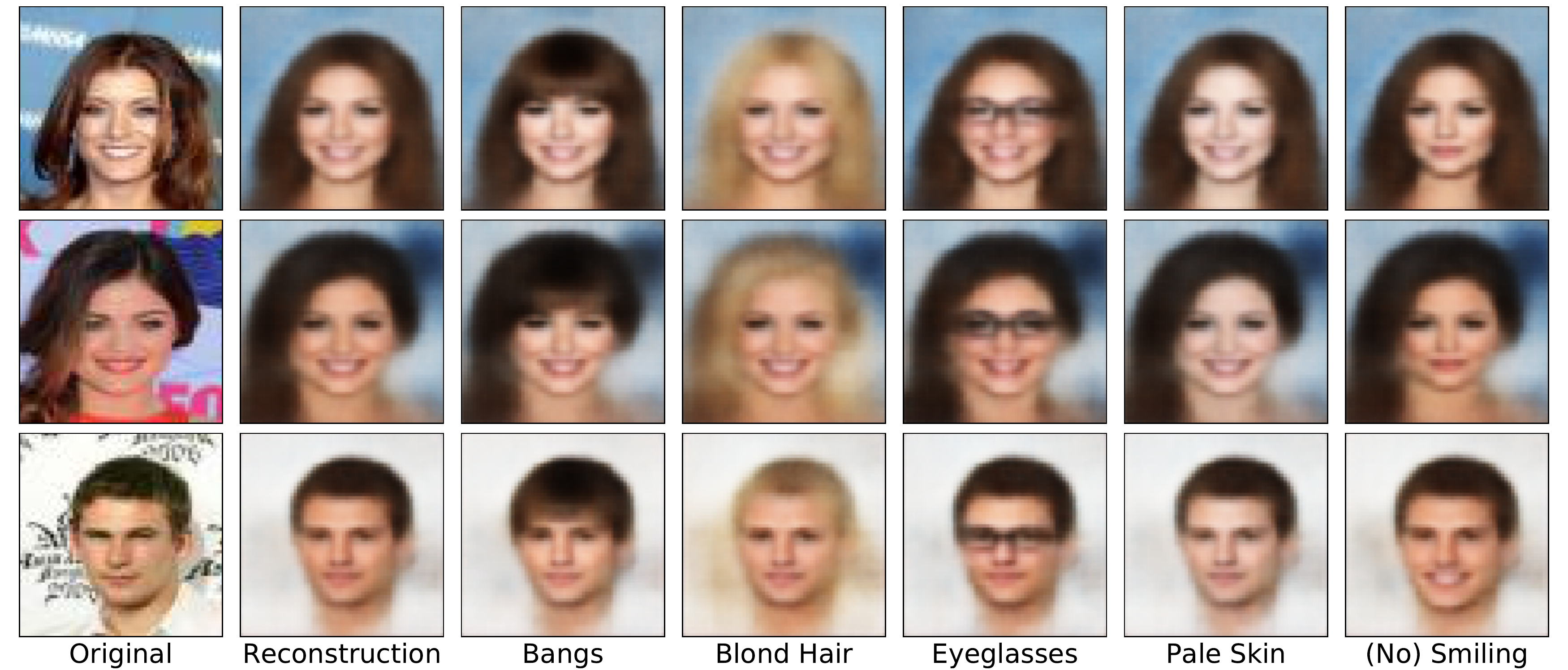}\hspace{0.2mm}
	\includegraphics[width=0.49\textwidth]{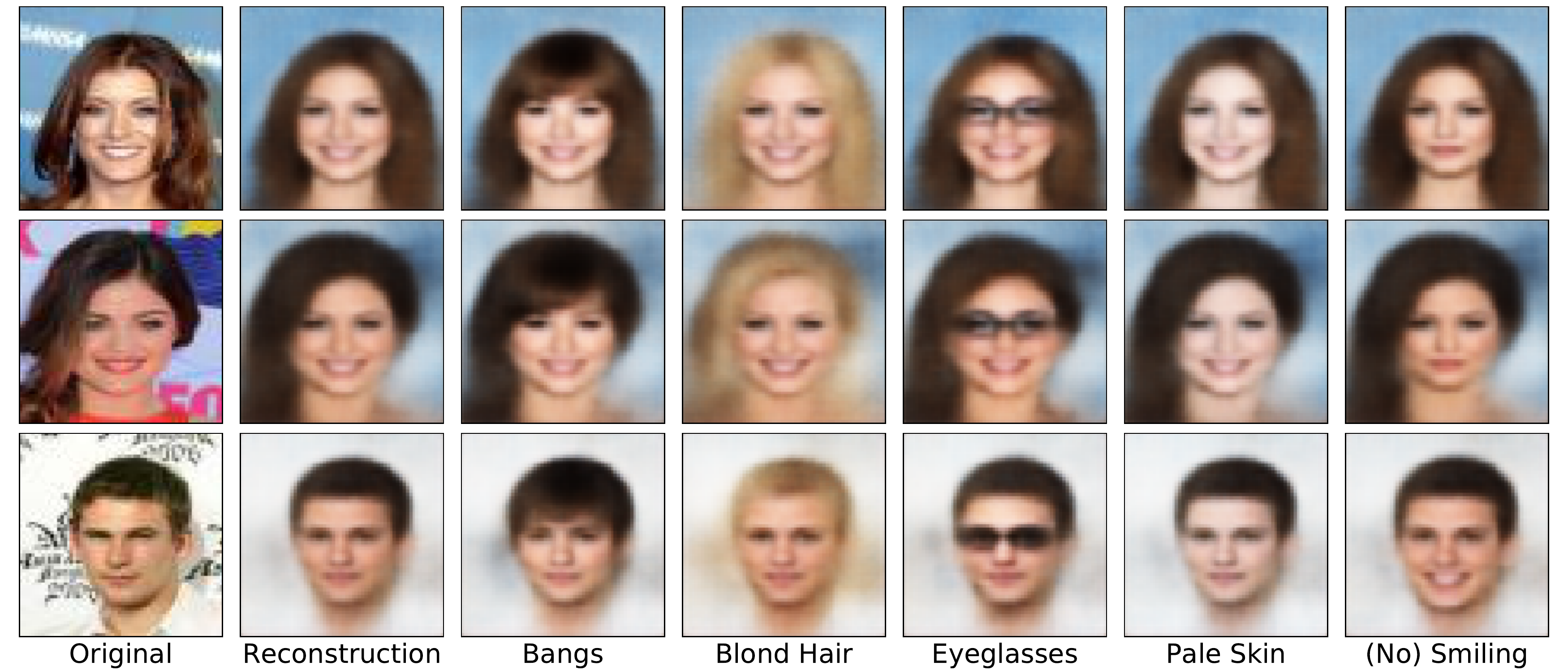}%
	\caption{Synthesized face images with the designated attributes. \textbf{Baselines}: \textbf{disAE} (top-left) and \textbf{disAE-XCov} (top-right). \textbf{Ours}: \textbf{mddAE-XCov} (bottom-left) and \textbf{mddAE-dCov} (bottom-right). In each panel, the image and attribute information are given below the third row.}\label{fig:generate_face_images}
\end{figure*}
\subsection{Verifying the Disentanglement Ability}\label{sec:disentangle_ability}
In the first experiment, we use the mddAE to generate new handwritten digits with the designated handwriting styles, such as boldface, italic, and broad shape. The image manipulation method has been introduced as the case 1 in Section \ref{sec:method_manipulate}. Due to the small size of MNIST image, here we don't employ the multi-scale SSIM to improve the generations' visual quality, but still obtain satisfactory results as shown in Figure \ref{fig:generate_digits}. We can see that the basic disAE, without any decorrelation regularizations but with a 2D $\mathbf{z}$-space, also learns a disentangled style representation from class label. However, this disentanglement ability is limited especially when considering the stroke thickness. By using a decorrelation regularization, both the disAE-XCov and the mddAE are able to generate novel digits with the same style as originals, demonstrating the disentanglement of style from class.

In the second experiment, we aim to synthesize new faces with the modified facial attributes while preserving the core identity. The manipulation method is described as the case 2 in Section \ref{sec:method_manipulate}. To alleviate the problem of blurry reconstruction, we incorporate the multi-scale SSIM into our model. As shown in Figure \ref{fig:generate_face_images}, for such a more complex dataset, the basic disAE fails to achieve disentanglement, and the identity information is easily destroyed in images generated from the disAE-XCov (e.g., the effects of adding blond hair). By contrast, our two mddAE models exhibit a remarkable ability to disentangle facial attributes from identity.

We also utilize three well-known image quality assessment indexes, namely Root-Mean-Square Error (RMSE), Peak Signal-to-Noise Ratio (PSNR), and SSIM mentioned above to evaluate the reconstructions' quality. As we can see from Table \ref{tab:recons_quality}, when using a discriminator or regularization, the reconstruction ability of disAE and disAE-XCov has an obvious degradation compared with their original counterparts. We believe this degradation comes from the function decomposition of the middle-layer representation in AE, which originally only focuses on reconstruction. However, with the structural dissimilarity as an auxiliary reconstruction loss, our model reconstructs images as accurately as AE, even achieving the best on SSIM. These results demonstrate that, the mddAE is adequate to learn disentangled representations without destroying the reconstruction performance. The same conclusion can also be found by comparing reconstructions and generations in Figure \ref{fig:generate_face_images} and \ref{fig:generate_face_images_diff_degrees}.
\begin{table}[ht]
	\renewcommand{\arraystretch}{1.3}
    \small
	\centering
	\caption{Reconstruction quality on the CelebA test set. Best two results are in bold.} \label{tab:recons_quality}
	\begin{tabular}{|l|*{3}{|c}|}
		\hline
		Model                         &RMSE                    &PSNR                    &SSIM            \\
		\hline\hline
		\multirow{2}{*}{AE}           &\textbf{0.0944}         &\textbf{20.7348}        &0.8817          \\
		                              &($\pm$0.0221)           &($\pm$2.0137)           &($\pm$0.0512)   \\ \hline
		\multirow{2}{*}{disAE}        &0.1144                  &19.0541                 &0.8457          \\
		                              &($\pm$0.0259)  		   &($\pm$1.9640)           &($\pm$0.0656)   \\ \hline
		\multirow{2}{*}{disAE-XCov}   &0.1173          		   &18.8098                 &0.8290          \\
									  &($\pm$0.0251)  		   &($\pm$1.8543)           &($\pm$0.0704)   \\ \hline
		\multirow{2}{*}{mddAE-XCov}   &\textbf{0.1066}         &\textbf{19.6596}        &\textbf{0.8914} \\
									  &($\pm$0.0237)  		   &($\pm$1.9259)           &($\pm$0.0424)   \\ \hline
		\multirow{2}{*}{mddAE-dCov}   &0.1078                  &19.5483                 &\textbf{0.8893} \\
									  &($\pm$0.0235)  		   &($\pm$1.8859)           &($\pm$0.0428)   \\
		\hline
	\end{tabular}
\end{table}

\subsection{Controllable Disentanglement}\label{sec:disentangle_degree}
\begin{figure*}
	\centering
	\includegraphics[width=0.33\textwidth]{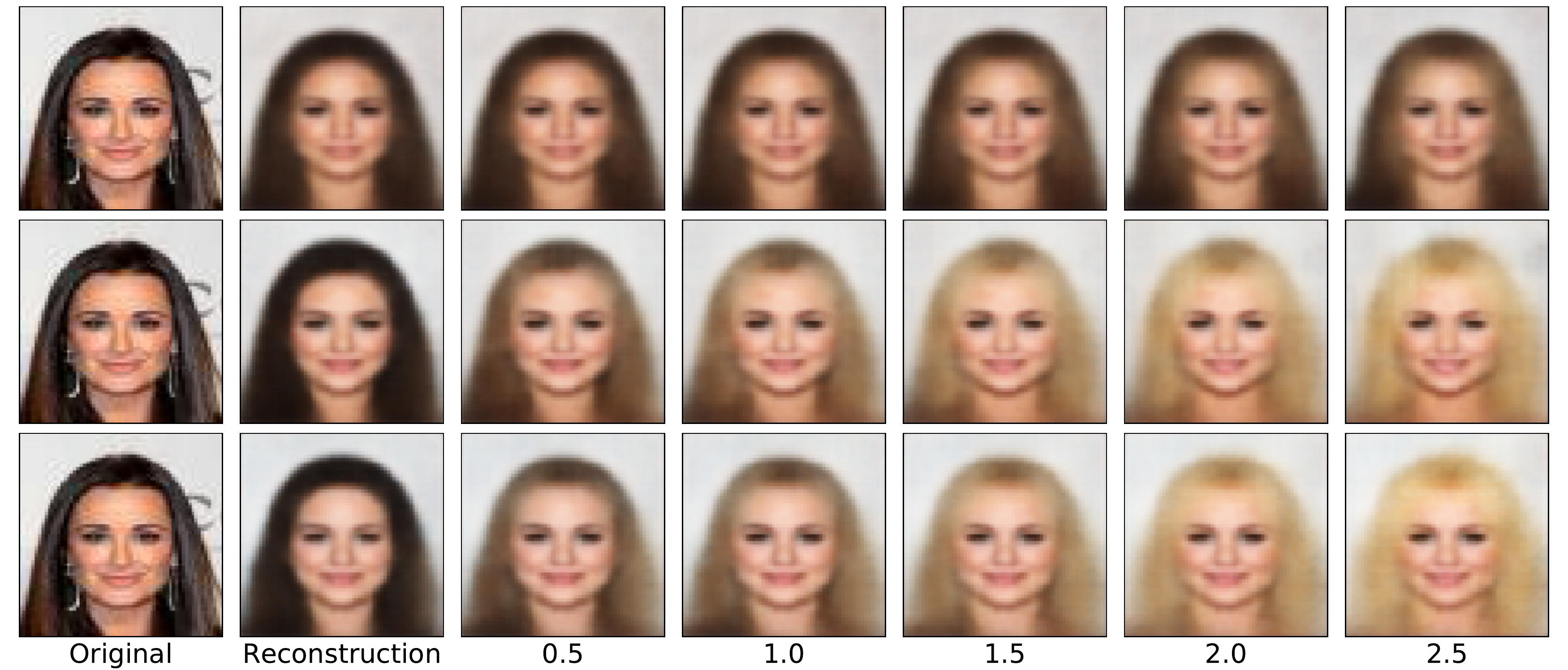}
	\includegraphics[width=0.33\textwidth]{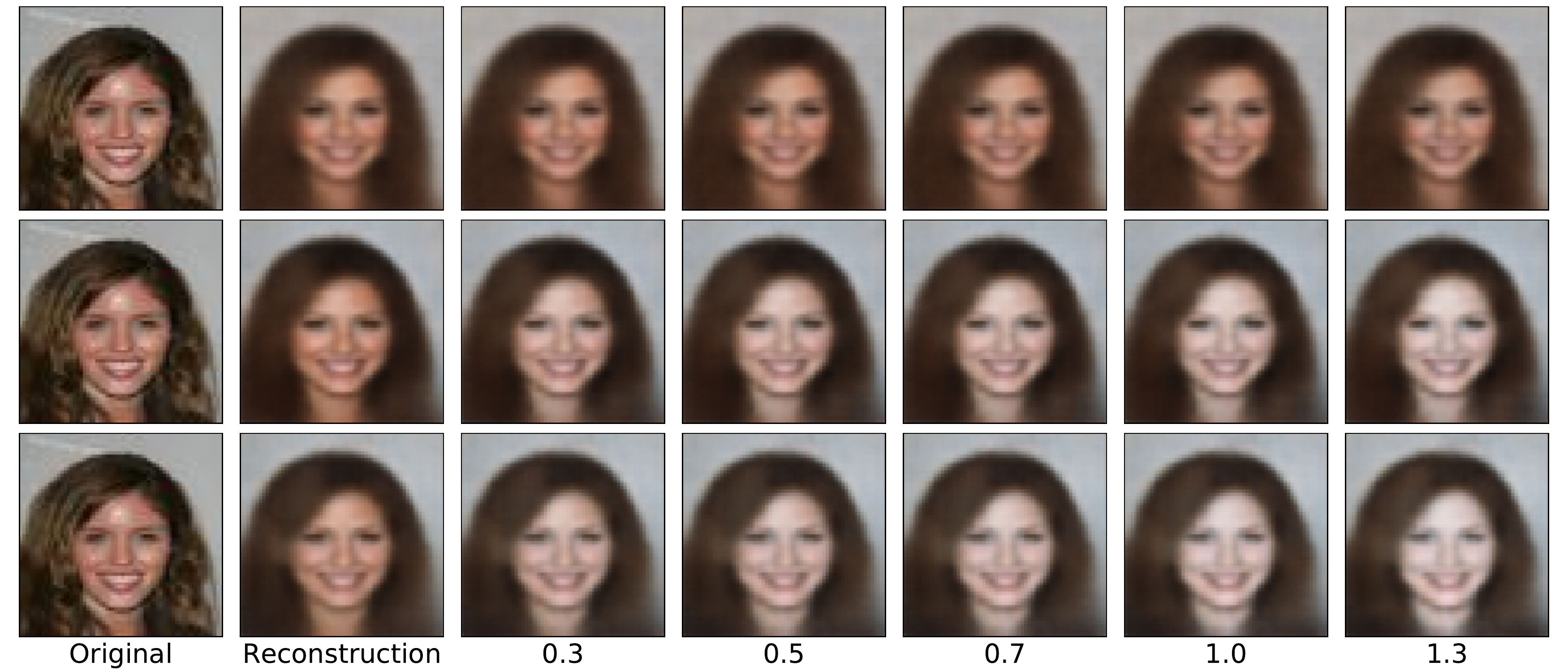}
	\includegraphics[width=0.33\textwidth]{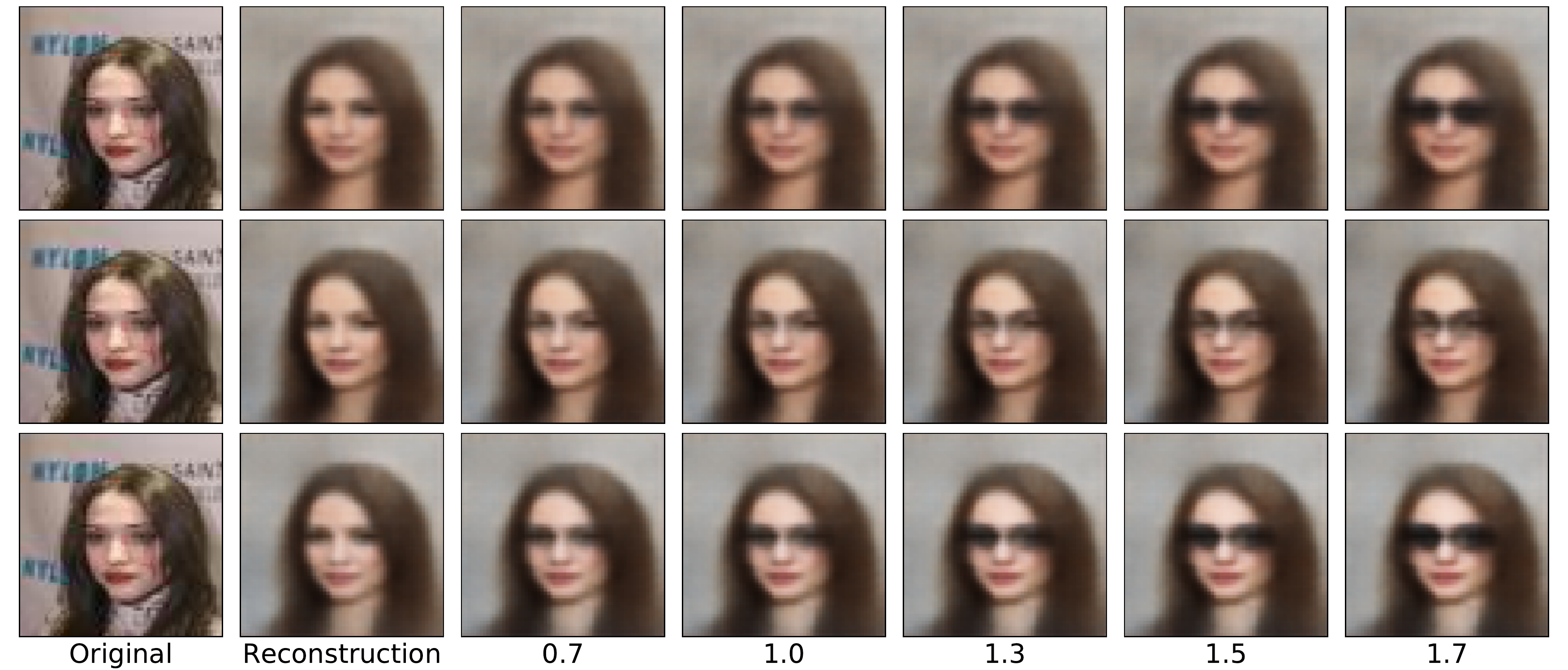}\\\vspace{0.5mm}
	\includegraphics[width=0.33\textwidth]{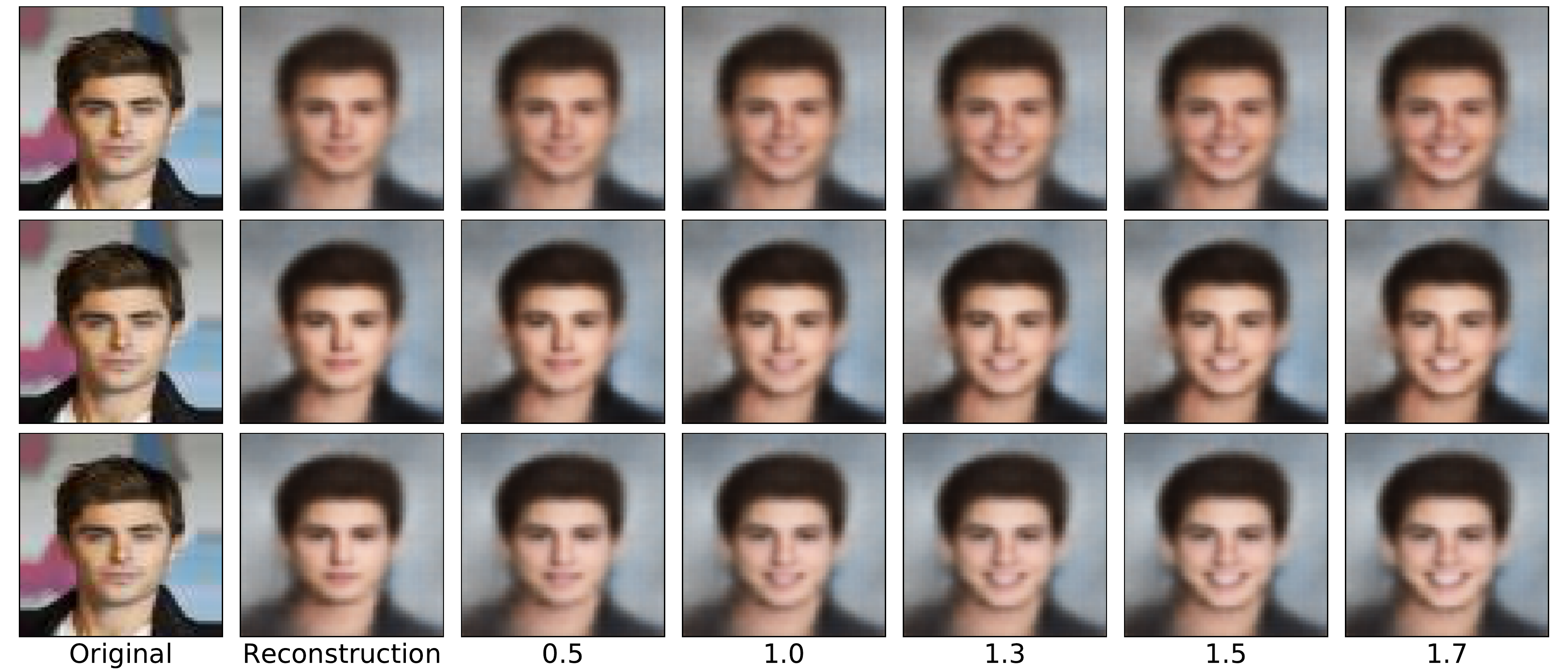}
	\includegraphics[width=0.33\textwidth]{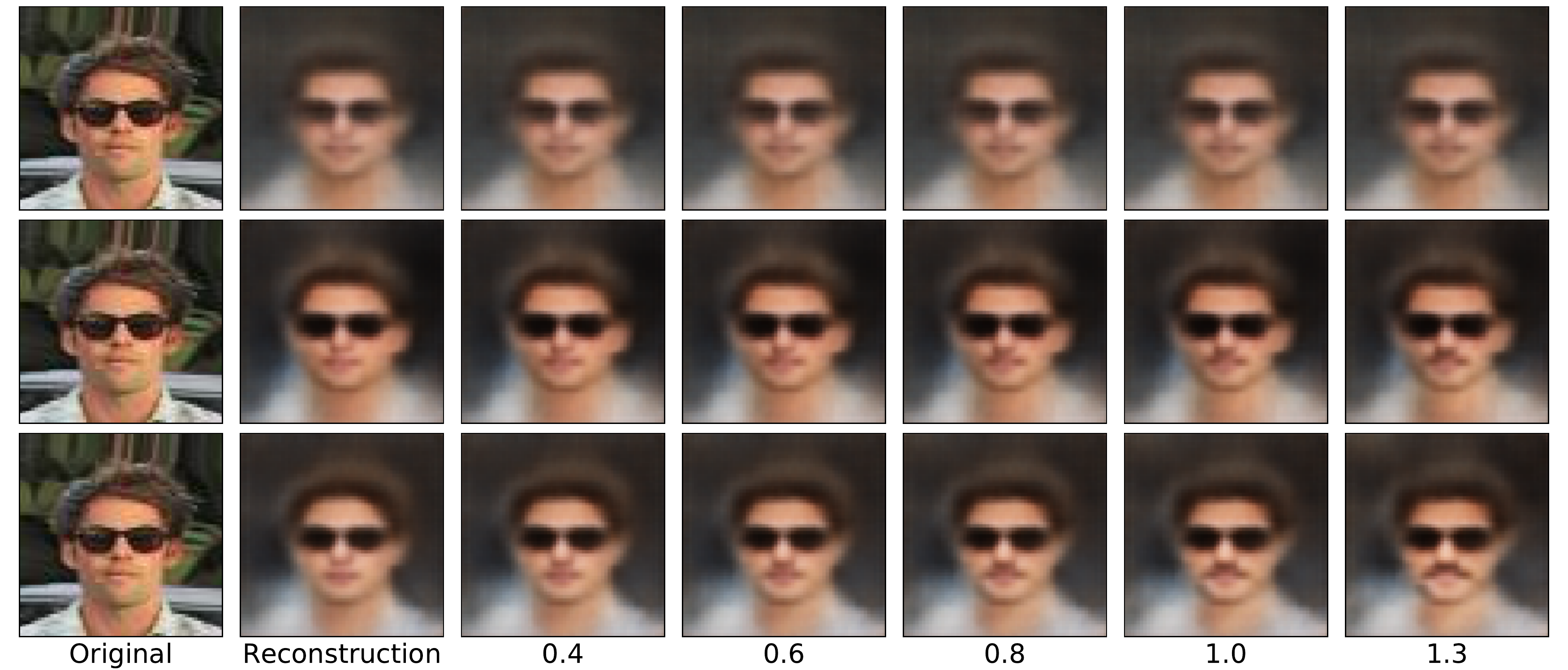}
	\includegraphics[width=0.33\textwidth]{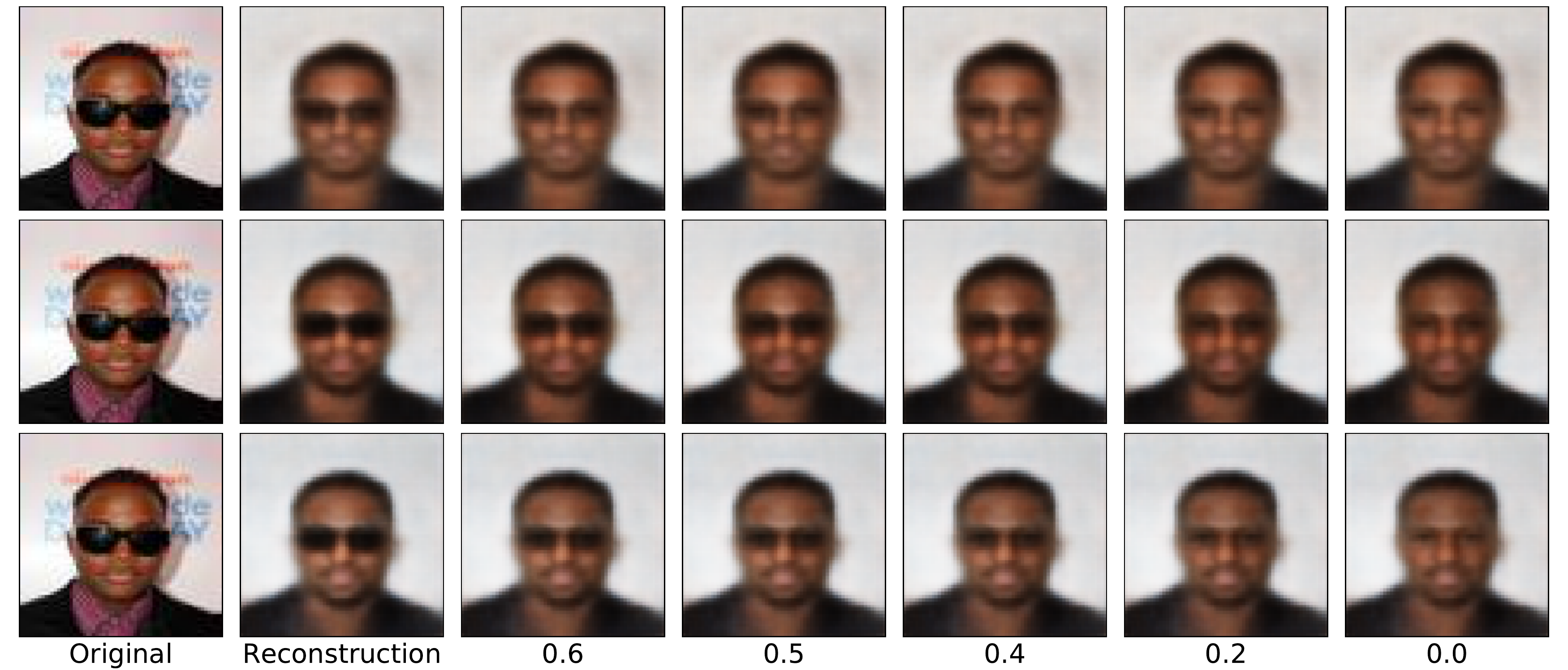}%
	\caption{Synthesized face images with various attribute intensities. Top row from left to right: \textbf{Blond Hair}, \textbf{Pale Skin}, \textbf{Eyeglasses}. Bottom row from left to right: \textbf{Smiling}, \textbf{Mustache}, \textbf{No Eyeglasses}. In each panel, the \textbf{first row} corresponds to the results for disAE-XCov, the \textbf{second row} for mddAE-XCov, the \textbf{third row} for mddAE-dCov, and the attribute intensity values are given below the third row. Zoom in for a better view.}\label{fig:generate_face_images_diff_degrees}
\end{figure*}
In this experiment, we \emph{qualitatively} compare the disentanglement strength of disAE-XCov and mddAE by manipulating face images with various attribute intensities. As can be seen from Figure \ref{fig:generate_face_images_diff_degrees}, while the disAE-XCov shows a visible difference between generated images on ``Eyeglasses'' and ``Smiling" attributes, it cannot generate faces with obvious difference on ``Blond Hair", ``Pale Skin", and ``Mustache" attributes, indicating a limited ability to control disentanglement. Our models, on the contrary, consistently generate distinguishable face fantasies across all compared attributes and variation degrees. These results illustrate that learning with the soft target representation enables the model to control disentanglement during image editing.

\subsection{Comparing Disentanglement Strength by Classification}
To further analyze the difference between XCov and dCov for disentanglement, we design an evaluation protocol to \emph{quantitatively} compare the disentanglement strength of the mddAE-XCov and the mddAE-dCov, which includes the following four steps.
\begin{itemize}
  \item First, divide the training set into two subsets: the first subset consists of images with the designated attribute, the second one not.
  \item Second, train a two-class classifier on these two subsets.
  \item Third,  in the test set, select all images that do not contain the designated attribute, then feed them to the disentanglement model to generate their counterparts with the designated attribute and intensity.
  \item Finally, employ the classifier trained in the second step to classify those images synthesized in the third step, obtaining a classification error rate as the evaluation index.
\end{itemize}

\begin{figure}[ht]
	\centering
	\includegraphics[width=0.47\textwidth]{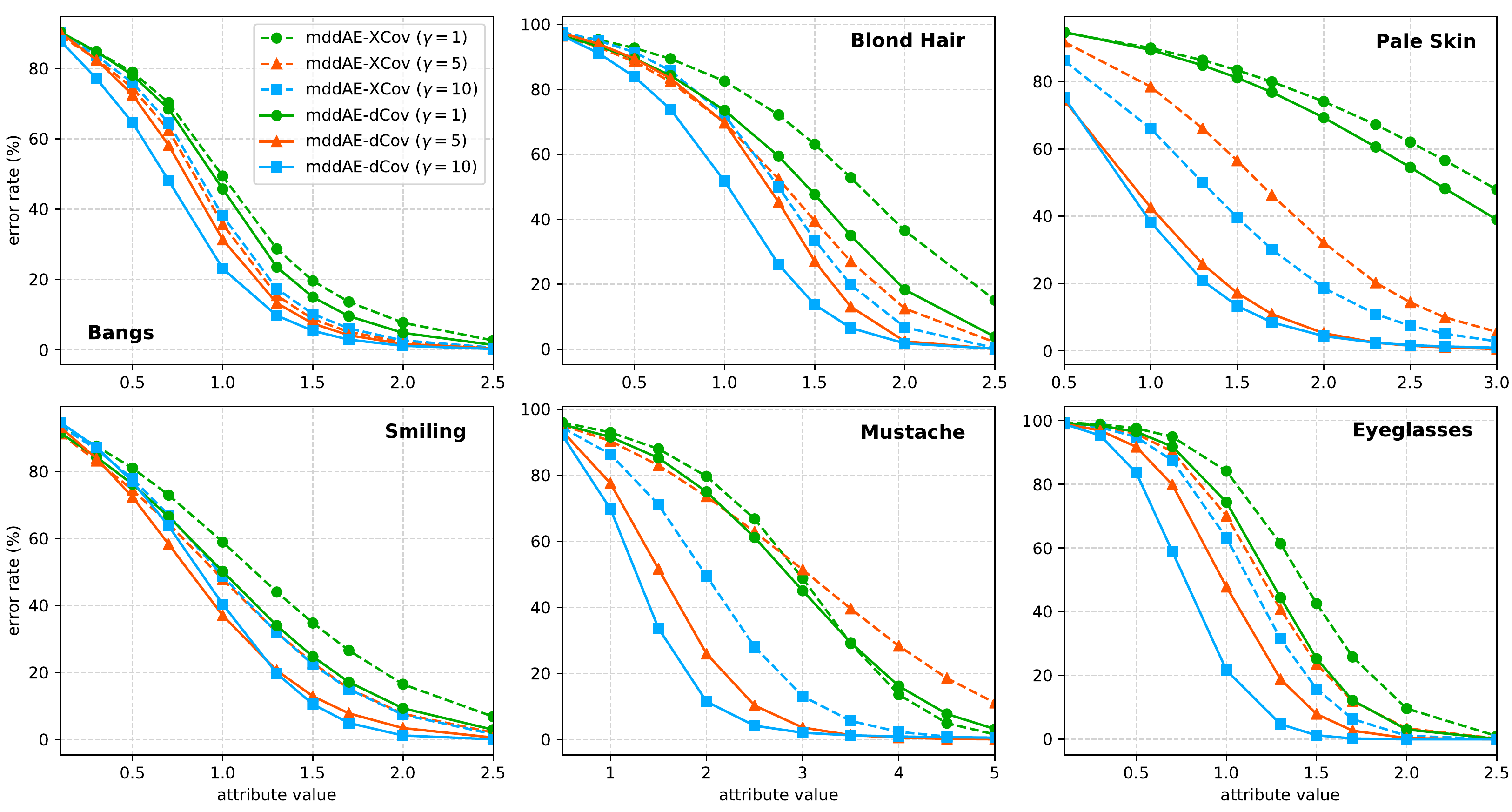}
	\caption{Attribute classification of synthesized face images by mddAE-XCov and mddAE-dCov. For each attribute, a linear SVM is ran for 5 times and we report the average performance. The bigger version of these figures and more results are provided in the supplementary material.}\label{fig:compare_disentangle_strength_by_class}
\end{figure}

The evaluation protocol is based on the hypothesis that the classifier is well-trained, and thus lower error rate means it's easier to perceive the designated attribute in synthesized images. For each attribute, we train a linear Support Vector Machine (SVM) as the two-class classifier to perform attribute classification task. The main evaluation results are depicted in Figure \ref{fig:compare_disentangle_strength_by_class}. We find that the classification error rates consistently decrease as increasing the attribute values, which implies that bigger attribute values could result in synthesized images with the more distinct attributes. In addition, with the same network architecture and the same regularization parameter $\gamma$, the classification performance corresponding to mddAE-dCov is superior to that of mddAE-XCov. We attribute this performance gap to the strong disentanglement ability of the dCov, that is, minimizing dCov encourages independence between random variables, rather than the non-correlation revealed in minimizing XCov.

\section{Conclusion}
We proposed a simple yet effective model that can learn disentangled representations, and can also manage the disentanglement degrees at image editing time. Briefly, a distance covariance based decorrelation regularization was devised to facilitate disentanglement, and the soft target representation was exploited to control how much a specific attribute is perceivable in the generated image. In addition, we also designed a classification protocol to evaluate the disentanglement strength of our model. Experimental results demonstrate that our model is able to generate new digits with various handwriting styles, and also synthesize new faces with the designated attributes and attribute intensities.

\section{Acknowledgments}
This work is supported in part by the National Natural Science Foundation of China (61572393, 11671317, 61877049), and in part by the China Scholarship Council. J. Zhang is supported by the National Key Research and Development Program of China (2018YFC0809001). The authors also thank Microsoft Azure for computing resources.

\bibliographystyle{aaai}
\bibliography{ref}

\end{document}